\documentclass{article}
\usepackage{ijcai19}
\usepackage{bm}
\usepackage{amsmath}
\usepackage{amsthm}
\usepackage{amssymb,amsfonts}
\usepackage{dsfont}

\newcommand{\argmin}{\operatornamewithlimits{arg\,min}}

\newtheorem{theorem}{Theorem}

\newtheorem{definition}{Definition}

\newtheorem{remark}{Remark}

\usepackage{times}
\usepackage{soul}
\usepackage{url}
\usepackage[hidelinks]{hyperref}
\usepackage[utf8]{inputenc}
\usepackage[small]{caption}
\usepackage{graphicx}
\usepackage{amsmath}
\usepackage{booktabs}
\usepackage{algorithm}
\usepackage{algorithmic}

\title{Efficient Cross-Validation for Semi-Supervised Learning}

\author{
Yong Liu$^1$
\and
Jian Li$^{12}$\and
Guangjun Wu$^1$\and
Lizhong Ding$^5$\And
Weiping Wang$^134$
\affiliations
$^1$Institute of Information Engineering, Chinese Academy of Sciences\\
$^2$School of Cyber Security, University of Chinese Academy of Sciencesn\\
$^3$National Engineering Research Center for Information Security\\
$^4$National Engineering Laboratory for Information Security Technology\\
$^5$Inception Institute of Artificial Intelligence (IIAI), Abu Dhabi, UAE
\emails
\{liuyong,lijian9026,wuguangjun,wangweiping\}@iie.ac.cn\\
lizhong.ding@inceptioniai.org
}

\begin{document}

\maketitle

\begin{abstract}
  Manifold regularization,
  such as laplacian regularized least squares (LapRLS) and laplacian support vector machine (LapSVM),
has been widely used in semi-supervised learning,
and its performance greatly depends on the choice of some hyper-parameters.
Cross-validation (CV) is the most popular approach for selecting the optimal hyper-parameters,
but it has high complexity due to multiple times of learner training.
In this paper, we provide a method to approximate the CV for manifold regularization based on a notion of robust statistics,
called  Bouligand influence function (BIF).
We first  provide a strategy for approximating
the CV via the Taylor expansion of BIF.
Then, we show how to calculate the BIF for general loss function,
and further give the approximate CV criteria for model selection in manifold regularization.
The proposed approximate CV for manifold regularization requires training only once,
hence can significantly improve the efficiency of traditional CV.
Experimental results show that our approximate CV has no statistical discrepancy with the original one,
but much smaller time cost.
\end{abstract}

\section{Introduction}
Semi-supervised learning (SSL),
which exploits the prior knowledge from the unlabeled data to improve performance,
has attracted considerable attention in recent years.
Manifold regularization,
such as
laplacian regularized least squares (LapRLS) \cite{belkin2006manifold},
laplacian support vector machine (LapSVM) \cite{sindhwani2005linear,belkin2006manifold},
has been widely used and very successful under the circumstance of a few of labeled
examples and amount of unlabeled examples.
The performance of these algorithms greatly depends on the choice of some hyper-parameters (such as kernel parameters,
regularization parameters and graph Laplacian parameters),
hence model selection is foundational to manifold regularization and is also a challenging problem in manifold regularization.

Cross-validation (CV) is a tried and tested approach
for selecting the optimal model  \cite{Josse2012}.
In $t$-fold CV,
data set is split into $t$ disjoint subset
of (approximately) equal size
and the learner machine is trained $t$ times,
each time leaving out one of subsets from training,
but using the omitted subset to compute the validation error.
The $t$-fold CV estimate is then simply the average
of the validation errors observed in each of the $t$ iterations, or folds.
Although $t$-fold CV is a commonly used approach for selecting the hyper-parameters, it requires training $t$ times,
making it disabled for large-scale model selection.

To address this problem, in this paper,
we present a novel approximate approach to CV for model selection of manifold regularization
based on a theoretical notion of Bouligand influence function (BIF) \cite{Christmann2008bif},
which requires training on the full data only once.
Specifically, we first introduce the notion of BIF,
and verify that it is the first derivative of an operator.
Then, we provide a strategy to approximate the CV via the first order approximation of Taylor expansion of BIF.
Finally, we propose a novel method to calculate the BIF for general loss function,
and further give two approximate CV criteria for LapRLS and LapSVM, respectively.
Experimental results on lots of datasets show that our approximate
CV has no statistical discrepancy with the original one,
but can significantly improve the efficiency.
This is the first attempt to use the notion of robust statistics to approximate
the general $t$-CV for model selection in SSL community.
\subsection*{Related Work}
In this subsection, we will introduce the related work
about approximate CV.
For kernel-based algorithms in supervised learning,
such as SVM, least square SVM (LSSVM), kernel ridge regression (KRR),
much work has been done to reduce the time complexity of leave-one-out CV,
see \cite{chapelle2002choosing,vapnik2000bounds,opper1999gaussian,Keerthi2002svm,liu2018fast}
for SVM,
\cite{Cawley2007pombrhp,Cawley2006} for LSSVM,
\cite{Cawley2004lssvm} for sparse LSSVM,
\cite{Debruyne2008msif,Debruyne2007phd} for KRR, etc.
\cite{Debruyne2008msif} proposed an approach to
approximating the leave-one-out CV based on the notion of influence function \cite{F.R.Hampel1986} for kernel-based regression.
Although there is much work on improving
the efficiency of the leave-one-out CV for kernel-based algorithms in supervised learning,
but little work focuses on the general $t$-CV.
\cite{Liu2014eacvkmbif} presented a strategy for approximating the general CV based
on the notion of Bouligand influence function (BIF) \cite{Christmann2008bif,Robinson1991aitcnf} for LSSVM and quadratic $\epsilon$-insensitive
support vector regression algorithms.
\cite{liu2019fast} extended the above method to various kernel-based algorithms,
and established an approximation theory of CV.
\cite{liu2018fast} further improved the efficiency of the computation of the BIF matrix for large scale problem.
However,
as far as we know,
how to approximate the general $t$-CV in semi-supervised learning community is still unknown.
In this paper, we will fill this gap.

The rest of the paper is organized as follows.
We introduce some notations and preliminaries in Section 2.
In Section 3, we present an approximate CV method via BIF.
In Section 4, we give the final model section criteria for LapRLS and LapSVM.
In Section 5, we empirically analyze the performance of our proposed approximate
CV. We end in Section 6 with a conclusion.
The proof is given in the last part.

\section{Notations and Preliminaries}
In semi-supervised learning,
we are given a small number of labeled examples $\mathcal{L}=\{(\mathbf x_i,y_i),i=1,\ldots,l\},$
and a large number of unlabeled examples $\mathcal{U}=\{\mathbf x_i,i=l+1,\ldots,l+u\},$
$l$ and $u$ denote the number of labeled and unlabeled points respectively.
Typically, $l\ll u$.
Let $\mathcal{S}=\mathcal{L}\cup\mathcal{U}$ be a set of $l+u$ examples, and
$\mathcal{T}=\mathcal{L}_\mathcal{X}\cup\mathcal{U}=\{\mathbf x_i,i=1,\ldots,l+u\}.$
Labeled examples are generated accordingly to the distribution $\mathbb{P}$ on $\mathcal{X}\times \mathcal{Y}$,
whereas unlabeled examples are drawn according to the marginal distribution $\mathbb{P}_\mathcal{X}$ of $\mathbb{P}$.
Labels are obtained from the conditional probability distribution $\mathbb{P}(y|\mathbf x)$.
For classification, $\mathcal{Y}=\{+1,-1\}$,
for regression, $\mathcal{Y}\in\mathbb{R}$.





Given a Mercer kernel $\kappa$ and its associated reproduce kernel Hilbert space (RKHS) $\mathcal{H}_\kappa$,
the manifold regularization can be written as
\begin{align*}
  f_{\mathbb{P}_\mathcal{S}}^\mathrm{MR}&=\argmin_{f\in\mathcal{H}_\kappa}\mathbb{E}_{(\mathbf x,y)\sim\mathbb{P}_\mathcal{S}}[\ell(y,f(\mathbf x))]+
  \gamma_A\|f\|_\kappa^2\\&~~~~~~~~~~~~~~~~~~+\gamma_I\mathbb{E}_{\mathbf x,\mathbf x'\sim\mathbb{P}_\mathcal{S}}\left[W_{ij}(f(\mathbf x)-f(\mathbf x'))^2\right],
  \\&=\argmin_{f\in\mathcal{H}_\kappa}\frac{1}{l}\sum_{i=1}^l \ell(y_i,f(\mathbf x_i))+\gamma_A\|f\|_\kappa^2+\frac{\gamma_I\bm f^\mathrm{T}\mathbf L_\mathcal{S}\bm f}{(l+u)^2},
\end{align*}
where $\mathbb{P}_\mathcal{S}$ is the empirical distribution, that is,
$\mathbb{P}_{\mathcal{S}}(z)=\frac{1}{l}$ when $z\in \mathcal{L}$,
$\mathbb{P}_{\mathcal{S}}(z)=\frac{1}{l+u}$ when $z\in \mathcal{T}$,
$\mathbb{P}_{\mathcal{S}}(z)=0$, otherwise,
$\ell:\mathcal{Y}\times\mathcal{Y}\rightarrow \mathbb{R}^+\cup\{0\}$ is a loss function,
$\|\cdot\|_\kappa$ is the norm in RKHS,
$\gamma_A$ is the weight of the norm of the function in the RKHS,
$\gamma_I$ is the weight of the norm of the function in the low dimensional manifold,
$\bm f=[f(\mathbf x_1),\ldots,f(\mathbf x_{l+u})]^\mathrm{T}$,
$\mathbf L_\mathcal{S} = \mathbf D -\mathbf W$ is the graph Laplacian matrix associated with $\mathcal{S}$,
where $\mathbf W$ is the affinity
matrix defining the similarity between any pair of samples of $\mathcal{S}$, and $\mathbf D$ is the
diagonal matrix with diagonal elements $\mathbf D_{ii}=\sum_{j=1}^{l+u}W_{ij}$.
Laplacian Regularized Least Squares (LapRLS) and Laplacian Support Vector Machine (LapSVM)
are two special cases of manifold regularized with different loss functions.
For LapRLS, $\ell$ is the square loss:
    $
      \ell(y,f(\mathbf x))=(f(\mathbf  x)-y)^2;
    $
For LapSVM, $\ell$ is the hinge loss:
    $
      \ell(y, f(\mathbf  x))=\max(0,1-yf(\mathbf x)).
    $

For the graph construction, a $k$NN graph is commonly used due to its simplicity and
effectiveness \cite{Melacci2011}.
In a $k$NN graph, the affinity matrix $\mathbf W$ is calculated as
$W_{ij}=\exp\big(-\frac{\|\mathbf x_i-\mathbf x_j\|^2}{2\sigma_w}\big),$
if $\mathbf x_i$ is the $k$ nearest neighbors of $\mathbf x_j$, and 0 otherwise.

Let $\{\mathcal{L}_i\}_{i=1}^t$ and $\{\mathcal{U}_i\}_{i=1}^t $ be a random equipartition of $\mathcal{L}$ and $\mathcal{U}$ into $t$ parts,
and
 $\mathcal{S}_i=\mathcal{L}_i\cup\mathcal{U}_i, \mathcal{T}_i={\mathcal{L}_i}_\mathcal{X}\cup \mathcal{U}_i, i=1,\ldots,t.$
For simplicity, assume that  $l$ and $u$ both mod $t$,
and hence, $|\mathcal{L}_i|=\frac{l}{t}=:m, |\mathcal{U}_i|=\frac{u}{t}=:n, i=1,\ldots,t.$
Let $\mathbb{P}_{\mathcal{S}\setminus \mathcal{S}_i}$ be the empirical distribution of $\mathcal{S}$ without the observations $\mathcal{S}_i$,
that is,
\begin{align}
\label{def-psdsi}
  \mathbb{P}_{\mathcal{S}\setminus\mathcal{S}_i}(z)=\left\{
  \begin{aligned}
    &\frac{1}{(t-1)m}, &&z\in \mathcal{L}\setminus\mathcal{L}_i,  \\
    &\frac{1}{(t-1)(m+n)}, &&z\in \mathcal{T}\setminus\mathcal{T}_i,\\
    &0, &&\text{otherwise}.
  \end{aligned}
  \right.
\end{align}
Let $f^\mathrm{MR}_{\mathbb{P}_{\mathcal{S}\setminus \mathcal{S}_i}}$ be the
hypothesis learned on all of the data excluding $\mathcal{S}_i$.
Then, the $t$-fold CV can be written as
 \begin{equation}
    \label{trad-CV}
    \text{$t$-CV}:=\sum_{i=1}^t\sum_{(\mathbf x_j,y_j)\in \mathcal{L}_i}
                    V\left(y_j,f^\mathrm{MR}_{\mathbb{P}_{\mathcal{S}\backslash \mathcal{S}_i}}(\mathbf  x_j)\right).
  \end{equation}
  From the above definition of $t$-CV,
  one can see that we need train the learning algorithm $t$ times to obtain $t$-CV.
  Thus, the time complexity of $t$-CV is $\mathcal{O}\big(\frac{(t-1)^3(l+u)^3}{t^2}\big),$
  which is computationally expensive.
  \begin{remark}
  The loss function $V$ adopted in the $t$-CV of \eqref{trad-CV} is not should to be the same as the learning machine.
  In this paper,  $V$ is the 0-1 loss for classification,
   $V$ is the square loss for regression.
\end{remark}
  \section{Approximate CV with BIF}
  In this section, we will first introduce the notion of Bouligand influence function (BIF),
  and further show how to use BIF to
  approximate $t$-CV in manifold regularization.
\subsection{Bouligand Influence Function}
\begin{definition}\cite{Christmann2008bif}
Let $\mathbb{P}$ be a distribution and $f^\mathrm{MR}$ be an operator $f^\mathrm{MR}:\mathbb{P}\rightarrow f^\mathrm{MR}_{\mathbb{\mathbb{P}}}$,
  then the \textbf{Bouligand influence  function (BIF)} of $f^\mathrm{MR}$ at $\mathbb{P}$ in the direction of a distribution $\mathbb{Q}\not=\mathbb{P}$ is defined as
  \begin{align*}
    \mathrm{BIF}(\mathbb{Q};f^{\mathrm{MR}},\mathbb{P})=\lim_{\epsilon\rightarrow 0}\frac{f^{\mathrm{MR}}_{(1-\epsilon)\mathbb{P}+
    \epsilon \mathbb{Q}}-f^{\mathrm{MR}}_{\mathbb{\mathbb{P}}}}{\epsilon}.
  \end{align*}
\end{definition}
One can see that BIF is used to measure the impact of an infinitesimal small amount of
contamination of the original distribution $\mathbb{P}$ in the direction of $\mathbb{Q}$ on the quantity of $f^\mathrm{MR}_\mathbb{P}$.
The notion of influence function \cite{F.R.Hampel1986}
popularly used in the field of robust statistics
is a special case of BIF
when setting $\mathbb{Q}$ to be the Dirac distribution $\delta_z$ at a point $z$.
BIF is a generalized notion of influence function.

Denote $\mathbb{P}_{\epsilon,\mathbb{Q}}=(1-\epsilon)\mathbb{P}+\epsilon \mathbb{Q}$.
Note that
the derivative of $f^\mathrm{MR}_{\mathbb{P}_{\epsilon,\mathbb{Q}}}$ at $\epsilon$ can be written as
$
  \lim_{\Delta\epsilon\rightarrow 0}
  \frac{f^\mathrm{MR}_{\mathbb{P}_{\epsilon+\Delta \epsilon,\mathbb{Q}}}-f^\mathrm{MR}_{\mathbb{P}_{\epsilon,\mathbb{Q}}}}{\Delta \epsilon}.
$
If setting $\epsilon=0$,
one can see that
$
  \lim_{\Delta\epsilon\rightarrow 0}
  \frac{f^\mathrm{MR}_{\mathbb{P}_{\Delta \epsilon,\mathbb{Q}}}-f^\mathrm{MR}_{\mathbb{P}}}{\Delta \epsilon}=\mathrm{BIF}(\mathbb{Q};f^\mathrm{MR},\mathbb{P}).
$
So, $\mathrm{BIF}(\mathbb{Q};f^\mathrm{MR},\mathbb{P})$ is the first order derivative of
$f^\mathrm{MR}_{\mathbb{P}_{\epsilon,\mathbb{Q}}}$ at $\epsilon=0$.
Thus, if BIF exists, the following  Taylor expansion holds:
\begin{align}
  \label{eq-taylor-Tp}
  f^{\mathrm{MR}}_{\mathbb{P}_{\epsilon,\mathbb{Q}}} \approx f^{\mathrm{MR}}_{\mathbb{\mathbb{P}}}+\epsilon\mathrm{BIF}(\mathbb{Q};f^{\mathrm{MR}},\mathbb{P}).
\end{align}

Note that
\begin{align*}
  \mathbb{\mathbb{P}}_{\mathcal{S}\backslash \mathcal{S}_i}=
  \Big(1-\Big(\frac{-1}{t-1}\Big)\Big)\mathbb{\mathbb{P}}_\mathcal{S}+
  \frac{-1}{t-1}\mathbb{P}_{\mathcal{S}_i},
\end{align*}
where $\mathbb{P}_{\mathcal{S}\setminus \mathcal{S}_i}$ is the empirical distribution of $\mathcal{S}$
without the observations $\mathcal{S}_i$ defined in \eqref{def-psdsi},
$\mathbb{P}_{\mathcal{S}_i}$
is the sample distribution corresponding to $\mathcal{S}_i$,
\begin{align*}
  \mathbb{P}_{\mathcal{S}_i}(z)=\left\{
  \begin{aligned}
    &\frac{1}{m}, &&z\in \mathcal{L}_i, \\
    &\frac{1}{m+n}, &&z\in \mathcal{T}_i, \\
    &0, &&\text{otherwise}.
  \end{aligned}
  \right.
\end{align*}
Thus, if taking
$
  \mathbb{Q}=\mathbb{P}_{\mathcal{S}_i}, \epsilon=\frac{-1}{t-1},
  \mathbb{P}_{\epsilon,\mathbb{Q}}=\mathbb{\mathbb{P}}_{\mathcal{S}\setminus \mathcal{S}_i},\mathbb{P}=\mathbb{\mathbb{P}}_\mathcal{S},
$
Equation (\ref{eq-taylor-Tp}) gives
\begin{align}
\label{Bk+1-LSSVM}
      f^{\mathrm{MR}}_{\mathbb{\mathbb{P}}_{\mathcal{S}\backslash \mathcal{S}_i}} \approx
      f^{\mathrm{MR}}_{\mathbb{\mathbb{P}}_\mathcal{S}}+\frac{1}{1-t}\mathrm{BIF}(\mathbb{P}_{\mathcal{S}_i};f^{\mathrm{MR}},\mathbb{\mathbb{P}}_\mathcal{S}).
\end{align}
Thus, the approximation of $t$-CV can be written as $t\text{-}\mathrm{BIF}:=$
\begin{align*}
 \sum_{i=1}^t\sum_{(\mathbf x_j,y_j)\in \mathcal{L}_i}
                    V\Big(y_j,f^{\mathrm{MR}}_{\mathbb{\mathbb{P}}_\mathcal{S}}(\mathbf x_j)+
                    \frac{\mathrm{BIF}(\mathbb{P}_{\mathcal{S}_i};f^{\mathrm{MR}},\mathbb{\mathbb{P}}_\mathcal{S})(\mathbf x_j)}{1-t}\Big).
\end{align*}
Note that $t\text{-}\mathrm{BIF}$ only depends on the calculation of  $f^{\mathrm{MR}}_{\mathbb{\mathbb{P}}_\mathcal{S}}$
 and $\mathrm{BIF}(\mathbb{P}_{\mathcal{S}_i};f^{\mathrm{MR}},\mathbb{\mathbb{P}}_\mathcal{S})$.
 Thus, if given the $\mathrm{BIF}(\mathbb{P}_{\mathcal{S}_i};f^{\mathrm{MR}},\mathbb{\mathbb{P}}_\mathcal{S})$,
 we need to train the algorithm only once on the full data set $\mathcal{S}$
 to obtain $f^\mathrm{MR}_{\mathbb{\mathbb{P}}_\mathcal{S}}$
 for approximating the
 $f^\mathrm{MR}_{\mathbb{\mathbb{P}}_{\mathcal{S}\backslash \mathcal{S}_i}}$, $i=1,\ldots,t$.
%
\subsection{The Compution of BIF}
Let $\mathbf K_{\mathcal{S}\mathcal{S}}$ be the $(l+u)\times(l+u)$
kernel matrix with $[\mathbf K_{\mathcal{S}\mathcal{S}}]_{jk}=\kappa(\mathbf x_j,\mathbf x_k)$,
$\mathbf K_{\mathcal{S}\mathcal{S}_i}$ the $(l+u)\times (m+n)$ kernel matrix
with $[\mathbf K_{\mathcal{S}\mathcal{S}_i}]_{jk}=\kappa(\mathbf x_j,\mathbf x_k)$, $\mathbf x_j\in \mathcal{T},\mathbf x_k\in \mathcal{T}_i$,
$\ell'(\cdot,\cdot)$ and $\ell''(\cdot,\cdot)$ be the first and second derivative of $\ell(\cdot,\cdot)$
with respect to the second variable respectively.

\begin{theorem}
\label{the-bif}
  Let $\mathbf{B}$ be the $(l+u)\times t$ BIF matrix with
  $[\mathbf{B}]_{j,i}=\mathrm{BIF}(\mathbb{P}_{\mathcal{S}_i};f^{\mathrm{MR}},\mathbb{\mathbb{P}}_\mathcal{S})(\mathbf x_j),$
  and
  $$\mathbf H=\frac{1}{l}\mathbf J_\mathcal{S}\mathbf K_{\mathcal{S}\mathcal{S}}\mathbf F_\mathcal{S}+2\gamma_A\mathbf I
    +\frac{2\gamma_I}{(l+u)^2}\mathbf K_{\mathcal{S}\mathcal{S}}\mathbf L_\mathcal{S},$$
  then the $i$th column of $\mathbf B$ can be written as
  \begin{align*}
    \begin{aligned}
    \mathbf B_{\cdot,i}=\mathbf H^{-1} \left[-\frac{\mathbf K_{\mathcal{S}\mathcal{S}_i}\bm \mu_{\mathcal{S}_i}}{m}
    -2\gamma_A \bm f_{\mathbb{P}_\mathcal{S}}^\mathcal{S}
    -\frac{2\gamma_I\mathbf K_{\mathcal{S}\mathcal{S}_i}\mathbf L_{\mathcal{S}_i}\bm f_{\mathbb{P}_\mathcal{S}}^{\mathcal{S}_i}}{(m+n)^2}\right],
  \end{aligned}
  \end{align*}
  where
  $\mathbf F_\mathcal{S}$ is an $(l+u)\times(l+u)$ diagonal matrix
with the first $l$ diagonal entries as $\ell''(y_j,f^\mathrm{MR}_{\mathbb{P}_{\mathcal{S}}}(\mathbf x_j))$ and the rest $0$,
$\bm \mu_{\mathcal{S}_i}$ is an $m+n$ vector with the fisrt $m$ entries as $\ell'(y^i_{j},f^\mathrm{MR}_{\mathbb{P}_{\mathcal{S}}}(\mathbf x^i_{j}))$,
$(\mathbf x^i_j,y^i_j)\in \mathcal{L}_i$, and the rest $0$,
$\mathbf L_{\mathcal{S}_i}$ is the graph Laplacian associated to $\mathcal{S}_i$,
 $\mathbf J_\mathcal{S}$ is an $(l+u)\times(l+u)$ diagonal matrix
with the first $l$ diagonal entries as $1$ and the rest $0$,
$\bm  f_{\mathbb{P}_\mathcal{S}}^\mathcal{S}=\big(f^\mathrm{MR}_{\mathbb{P}_\mathcal{S}}(\mathbf x_1),\ldots,
f^\mathrm{MR}_{\mathbb{P}_\mathcal{S}}(\mathbf x_{l+u})\big)^\mathrm{T}$,
$\bm  f_{\mathbb{P}_\mathcal{S}}^{\mathcal{S}_i}=\big(f^\mathrm{MR}_{\mathbb{P}_\mathcal{S}}(\mathbf x^i_1),
\ldots, f^\mathrm{MR}_{\mathbb{P}_\mathcal{S}}(\mathbf x^i_{m+n})\big)^\mathrm{T}$, $\mathbf x_j^i\in\mathcal{T}_i$.
\end{theorem}
The above theorem shows that if the first and second derivative of loss function $\ell$ exists,
the BIF can be obtained.
In the following, we will show how to calculate the BIF matrix for LapRLS and LapSVM, respectively.


\subsubsection{Laplacian Regularized Least Squares (LapRLS)}
Note that the loss function of LapRLS is the least square loss,
according to the definitions of $\mathbf F_\mathcal{S}$ and $\bm \mu_{\mathcal{S}_i}$ in Theorem \ref{the-bif},
it is easy to verity that
$\mathbf F_\mathcal{S}$
is the diagonal matrix with the first $l$ entries as 2 and the rest $0$,
$\bm \mu_{\mathcal{S}_i}$ is an $m+n$ vector with the fisrt $m$ entries as
 $2(f^\mathrm{MR}_{\mathbb{P}_{\mathcal{S}}}(\mathbf x^i_{j})-y^i_{j})$, $(\mathbf x_j^i,y_j)\in\mathcal{L}_i$, and the rest $0$.
Thus, from Theorem \ref{the-bif}, the BIF matrix of LapRLS can be written as
  \begin{align*}
    \begin{aligned}
    \mathbf B_{\cdot,i}=\mathbf H^{-1}\left[-\frac{\mathbf K_{\mathcal{S}\mathcal{S}_i}\bm \mu_{\mathcal{S}_i}}{m}
    -2\gamma_A \bm f_{\mathbb{P}_\mathcal{S}}^\mathcal{S}
    -\frac{2\gamma_I\mathbf K_{\mathcal{S}\mathcal{S}_i}\mathbf L_{\mathcal{S}_i}\bm f_{\mathbb{P}_\mathcal{S}}^{\mathcal{S}_i}}{(m+n)^2}\right],
  \end{aligned}
  \end{align*}
  where $\mathbf H=\frac{2\mathbf J_\mathcal{S}\mathbf K_{\mathcal{S}\mathcal{S}}\mathbf J_\mathcal{S}}{l}+2\gamma_A\mathbf I
    +\frac{2\gamma_I \mathbf K_{\mathcal{S}\mathcal{S}}\mathbf L_\mathcal{S}}{(l+u)^2}.$
\subsubsection{Laplacian Support Vector Machine (LapSVM)}
Since the hinge loss
$\ell(y, t)=\max(0,1-yt)$ is not differentiable,
but according to Theorem \ref{the-bif},
to obtain the BIF matrix, loss function should be differentiable.
Thus, we propose to use a differentiable approximation of it, inspired by the Huber loss:
\begin{align*}
 \ell(y,t)=
 \left\{
  \begin{aligned}
    &0                    &\text{ if }& yt>1+h,\\
    &\frac{(1+h-yt)^2}{4h}&\text{ if }& |1-yt|\leq h,\\
    &1-yt                 &\text{ if }& yt<1-h.
  \end{aligned}
  \right.
\end{align*}
Note that if $h\rightarrow 0$, the Huber loss converges to the hinge loss.
From the Huber loss, we know that
\begin{align*}
 &\ell'(y,t)=
 \left\{
  \begin{aligned}
    &0                    &\text{ if } & yt>1+h,\\
    &\frac{-y(1+h-yt)}{2h}&\text{ if } & |1-yt|\leq h,\\
    &-y                   &\text{ if } &yt<1-h,
  \end{aligned}
  \right.\\
 & \ell''(y,t)=
 \left\{
  \begin{aligned}
   &0            &\text{ if } & yt>1+h,\\
   &\frac{1}{2h} &\text{ if } & |1-yt|\leq h,\\
   &0            &\text{ if } & yt<1-h.
  \end{aligned}
  \right.
\end{align*}
\begin{table*}
\small
    \centering
    \caption{
 Test errors for
 classification and test mean square errors for regression.
 Our methods: \texttt{$t$-FBIF},
 compared methods: $t$-CV (\texttt{$t$-CV}),
 $t$=5,10,20.
 }
 \label{tab-test-classification}
    \begin{tabular*}{\linewidth}{@{\extracolsep{\fill}}crrrrrr}
    \toprule
    Classification     & \texttt{5-CV} &\texttt{5-BIF}   &\texttt{10-CV}   &\texttt{10-BIF}   &\texttt{20-CV} &\texttt{20-BIF}\\
    \midrule
       a1a       &20.33 $\pm$    3.22 &20.58    $\pm$ 2.05        &20.54 $\pm$ 2.16    &21.30 $\pm$ 2.29    &20.61 $\pm$    2.85    &21.53    $\pm$ 2.75\\
       a2a       &20.05    $\pm$ 1.45     &21.60    $\pm$ 0.61     &20.05    $\pm$ 1.45     &21.99    $\pm$ 0.17     &20.05    $\pm$ 1.45     &21.12    $\pm$ 1.06\\
       a3a       &18.51    $\pm$ 0.21    &19.91    $\pm$ 2.03    &18.44    $\pm$ 0.32    &19.50    $\pm$ 2.18    &19.18    $\pm$ 0.47    &20.42    $\pm$ 2.88\\
       fourclass &4.89    $\pm$ 2.79    &5.21  $\pm$ 2.36    &4.89 $\pm$    2.79    &5.21 $\pm$    2.36    &4.50 $\pm$    2.13    &5.12 $\pm$    2.59\\
       german    &33.00    $\pm$ 4.82    &33.22    $\pm$ 2.99    &33.00    $\pm$ 4.82    &33.66    $\pm$ 0.69    &32.45    $\pm$ 4.93    &33.73    $\pm$ 1.34 \\
       madelon     &45.65    $\pm$ 2.34  &46.36    $\pm$ 2.11        &45.65    $\pm$ 2.34 &47.14    $\pm$ 1.42        &46.48    $\pm$ 2.11    &46.76 $\pm$     1.92\\
       svmguide3 &20.55    $\pm$ 2.60    &20.29 $\pm$    4.24    &20.38 $\pm$ 2.90    &20.73 $\pm$ 4.46    &18.95 $\pm$    3.57    &19.75    $\pm$ 4.96\\
       splice    &29.57    $\pm$ 1.76    &30.44    $\pm$ 1.07    &29.57    $\pm$ 1.76    &30.01    $\pm$ 0.84    &29.13    $\pm$ 1.38 &29.57    $\pm$ 1.76    \\
       w1a       &3.14    $\pm$ 0.28  &3.18    $\pm$ 0.31        &3.14 $\pm$ 0.28 &3.32    $\pm$ 0.08        &3.05    $\pm$ 0.21 &3.32    $\pm$ 0.08    \\
       w2a       &3.04    $\pm$ 0.28    &3.07    $\pm$ 0.25    &3.04    $\pm$ 0.28    &3.68    $\pm$ 1.22    &3.04    $\pm$ 0.28    &3.04    $\pm$ 0.28\\
       w3a       &2.40    $\pm$ 0.31    &2.40    $\pm$ 0.31    &2.40    $\pm$ 0.31  &2.42    $\pm$ 0.27        &2.44    $\pm$ 0.29    &2.69    $\pm$ 0.45\\
        \midrule
       Regression     & \texttt{5-CV} &\texttt{5-BIF}   &\texttt{10-CV}   &\texttt{10-BIF}   &\texttt{20-CV} &\texttt{20-BIF}\\
       \midrule
      abalone  &8.71$\pm$0.81(e-04)    &8.71$\pm$0.81(e-04)    &8.57$\pm$0.75(e-04)    &8.57$\pm$0.75(e-04)    &8.52$\pm$0.64(e-04)    &8.52$\pm$0.64(e-04)\\
      bodfat   &9.23$\pm$5.96(e-02)    &9.23$\pm$5.96(e-02)    &9.31$\pm$5.46(e-02)    &9.31$\pm$5.46(e-02)    &9.15$\pm$5.67(e-02)    &9.15$\pm$5.67(e-02)\\
      cpusmall &2.94$\pm$0.40(e-02) &2.94$\pm$0.40(e-02)    &2.94$\pm$0.40(e-02)    &2.94$\pm$0.40(e-02)    &2.94$\pm$0.40(e-02)    &2.94$\pm$0.40(e-02)\\
      housing  &2.56$\pm$0.24(e-01) &2.63$\pm$0.19(e-01)    &2.56$\pm$0.24(e-01)    &2.63$\pm$0.19(e-01)        &2.54$\pm$0.18(e-01)    &2.58$\pm$0.57(e-01)\\
      mg       &1.16$\pm$0.06(e-01) &1.16$\pm$0.06(e-01)    &1.16$\pm$0.06(e-01)    &1.16$\pm$0.06(e-01)        &1.16$\pm$0.06(e-01)    &1.16$\pm$0.06(e-01)    \\
       mpg      &3.30$\pm$0.18(e-01)    &3.39$\pm$0.19(e-01)    &3.42$\pm$0.21(e-01)    &3.53$\pm$0.15(e-01)    &3.18$\pm$0.18(e-01)    &3.18$\pm$0.18(e-01)\\
      space-ga &8.31$\pm$0.18(e-02)    &8.31$\pm$0.18(e-02)        &8.32$\pm$0.18(e-02)    &8.32$\pm$0.18(e-02)    &8.32$\pm$0.18(e-02)    &8.32$\pm$0.18(e-02)    \\
    \bottomrule
 \end{tabular*}
 \vspace{-0.4cm}
\end{table*}

We say that $\mathbf x_i$ is a $support~vector$
if $|y_i(f^\mathrm{MR}_{\mathbb{\mathbb{P}}_\mathcal{S}}(\mathbf x_i))-1|< h.$
Let us reorder the training points
such that the first labeled $l_{\mathrm{sv}}$ points are support vectors.
From the definition of $\mathbf F_\mathcal{S},$ one can see that
$
  \mathbf F_\mathcal{S}=\frac{1}{2h}\mathbf I_{\mathrm{sv}},
$
where $\mathbf I_{\mathrm{sv}}$ is the $(l+u)\times (l+u)$
diagonal matrix with the first $l_\mathrm{sv}$ entries being 1 and the others 0,
$\bm \mu_{\mathcal{S}_i}$ is an $m+n$ vector with the first $m$ entries as
 $\ell'(y^i_{j}, f^\mathrm{MR}_{\mathbb{P}_{\mathcal{S}}}(\mathbf x^i_{j}))$, $(\mathbf x_j^i,y_j)\in\mathcal{L}_i$, and the rest $0$.
Thus, according to  Theorem \ref{the-bif},
the BIF matrix of LapSVM can be written as
\begin{align*}
   \mathbf B_{\cdot,i}=\mathbf H^{-1}\left[-\frac{\mathbf K_{\mathcal{S}\mathcal{S}_i}\bm \mu_{\mathcal{S}_i}}{m}
    -2\gamma_A \bm f_{\mathbb{P}_\mathcal{S}}^\mathcal{S}
    -\frac{2\gamma_I\mathbf K_{\mathcal{S}\mathcal{S}_i}\mathbf L_{\mathcal{S}_i}\bm f_{\mathbb{P}_\mathcal{S}}^{\mathcal{S}_i}}{(m+n)^2}\right],
\end{align*}
where $\mathbf H=\frac{\mathbf J_\mathcal{S}\mathbf K_{\mathcal{S}\mathcal{S}}\mathbf I_\mathrm{sv}}{2lh}+2\gamma_A\mathbf I
    +\frac{2\gamma_I \mathbf K_{\mathcal{S}\mathcal{S}}\mathbf L_\mathcal{S}}{(l+u)^2}.$

\begin{remark}
  In this paper, we only consider the use of square loss and hinge loss,
  but it is easy to extend our result to other loss functions,
  such as square hinge loss $\max(0,1-yt)^2$,
  logistic loss $\ln(1+\exp(-yt))$, and so on.
\end{remark}
\section{Model Selection}
According to the above discussion,
we know that
$$
 t\text{-}\mathrm{BIF}:=\sum_{i=1}^t\sum_{(\mathbf x_j,y_j)\in \mathcal{L}_i}
                    V\left(y_j,f^{\mathrm{MR}}_{\mathbb{\mathbb{P}}_\mathcal{S}}(\mathbf x_j)+
                    \frac{\mathbf B_{ji}}{1-t}\right)
$$
is an efficient approximation of CV for manifold regularization, which only need to training once.
However, to obtain $t$-BIF,
we need $\mathcal{O}\big((l+u)^3\big)$ to calculate $\mathbf H^{-1}$ to obtain the BIF matrix $\mathbf B$.

To accelerate the computation of the inversion of $\mathbf  H$,
we consider the use of the popular Nystr\"{o}m method.
Suppose we randomly sample $c$ columns of the matrix $\mathbf K_{\mathcal{S}\mathcal{S}}$  uniformly without replacement.
Let $\mathbf C$ be the $n\times c$ martix formed by theses columns,
 $\mathbf P$  the $c\times c$ matrix consisting of the intersection of these $c$ columns with the corresponding
$c$ rows of $\mathbf K_{\mathcal{S}\mathcal{S}}$.
Without loss of generality, we can rearrange the columns and rows of $\mathbf K_{\mathcal{S}\mathcal{S}}$
based on this sampling such that:
\begin{align*}
  \mathbf K_{\mathcal{S}\mathcal{S}}=\left(
   \begin{aligned}
           &\mathbf P, &&{{\mathbf K^\mathrm{T}_{\mathcal{S}_c\mathcal{S}}}} \\
           &{\mathbf K_{\mathcal{S}_c\mathcal{S}}}, &&\mathbf K_{(\mathcal{S}\backslash\mathcal{S}_c)(\mathcal{S}\backslash\mathcal{S}_c)}
         \end{aligned}
  \right),
\mathbf C=\left(\begin{array}{c}
           \mathbf P\\
           {\mathbf K_{\mathcal{S}_c\mathcal{S}_c}}
         \end{array}
  \right).
\end{align*}
The Nystr\"{o}m method uses $\mathbf P$ and $\mathbf C$ to construct an approximation
$\widetilde{\mathbf K}$ of $\mathbf K$ defined by:
\begin{align}
  \label{mathbfKplus}
  \widetilde{\mathbf K}_{\mathcal{S}\mathcal{S}}=\mathbf C\mathbf P^{+}\mathbf C^\mathrm{T}\approx \mathbf K_{\mathcal{S}\mathcal{S}},
\end{align}
where $\mathbf P^{+}$ is the Moore-Penrose generalized inverse of $\mathbf P$.

Denote $\mathbf T$ as
\begin{align*}
  \mathbf T=\left\{
  \begin{aligned}
    &\frac{2\mathbf J_\mathcal{S}\mathbf K_{\mathcal{S}\mathcal{S}}\mathbf J_\mathcal{S}}{l}+2\gamma_A\mathbf I, &&\text{for LapRLS}\\
    &\frac{\mathbf J_\mathcal{S}\mathbf K_{\mathcal{S}\mathcal{S}}\mathbf I_\mathrm{sv}}{2lh}+2\gamma_A\mathbf I, && \text{for LapSVM}
  \end{aligned}
  \right.,
\end{align*}
so $\mathbf H$ can be approximated by
$
  \tilde{\mathbf H}=\mathbf T
    +\frac{2\gamma_I}{(l+u)^2}\mathbf C\mathbf P^{+}(\mathbf C^\mathrm{T}\mathbf L_\mathcal{S}).
$

According to the Woodbury formula:
\begin{align*}
  \left(\mathbf A+\mathbf X\mathbf Y\mathbf Z\right)^{-1}=\mathbf A^{-1}
  -\mathbf A^{-1}\mathbf X(\mathbf Y^{-1}+\mathbf Z\mathbf A^{-1}\mathbf X)^{-1}\mathbf Z\mathbf A,
\end{align*}
it is easy to verity that $\tilde{\mathbf H}^{-1}=$
\begin{align*}
  \mathbf T^{-1}-\frac{2\gamma_I\mathbf T^{-1}\mathbf C}{(1+u)^2}
  \Big[\mathbf P+\frac{2\gamma_I\mathbf C^\mathrm{T}\mathbf L_\mathcal{S}\mathbf T^{-1}\mathbf C}{(1+u)^2}\Big]^{-1}\mathbf C^\mathrm{T}\mathbf L_\mathcal{S}\mathbf T,
\end{align*}
where
\begin{align*}
  \mathbf T^{-1}
  &=
  \left\{
  \begin{aligned}
       & \left[
                \begin{aligned}
                  &\left(\frac{\mathbf K_{\mathcal{L}\mathcal{L}}}{2l}+2\gamma_A\mathbf I\right)^{-1} &&\bm 0\\
                  &\bm 0^\mathrm{T}                                  &&\frac{\mathbf I}{2\gamma_A}
                \end{aligned}
               \right],
      &&\text{LapRLS}\\
 &\left[
 \begin{aligned}
                  &\left(\frac{\mathbf K_{\mathcal{L}_\mathrm{sv}\mathcal{L}_\mathrm{sv}}}{2lh}+2\gamma_A\mathbf I\right)^{-1} &&\bm 0\\
                  &\bm 0^\mathrm{T}                                  &&\frac{\mathbf I}{2\gamma_A}
                \end{aligned}
                \right]. &&\text{LapSVM}
  \end{aligned}
  \right.
\end{align*}
Note that $\mathbf P+\frac{2\gamma_I}{(1+u)^2}\mathbf C^\mathrm{T}\mathbf L_\mathcal{S}\mathbf T^{-1}\mathbf C\in \mathbb{R}^{c\times c},$
and the time complexity of $\mathbf T^{-1}$ is $\mathcal{O}(l^3)$ for LapRLS and $\mathcal{O}(l^3_\mathrm{sv})$ for LapSVM,
so we only need $\mathcal{O}(l^3+c^3+(l+u)c^2)$ and $\mathcal{O}(l_\mathrm{sv}^3+c^3+(l+u)c^2)$
to compute the $\tilde{\mathbf H}^{-1}$ for LapRLS and LapSVM, respectively.

Therefore, in this paper,
we finally consider the use of the following fast $t$-fold CV for model selection:
  \begin{equation}
  \label{appro-bif-finally}
  t\text{-}\mathrm{BIF}:=\sum_{i=1}^t\sum_{z_j\in \mathcal{L}_i}
                    V\Big(y_j, f^\mathrm{MR}_{\mathbb{\mathbb{P}}_{\mathcal{S}}}(\mathbf  x_j)+
      \frac{[\widetilde{\mathbf B}^\mathrm{SVM}]_{ji}}{1-t}\Big),
  \end{equation}
where $\widetilde{\mathbf B}$ is the approximation of $\mathbf B$ with $\widetilde{\mathbf H}$
replace of  $\mathbf{H}$.
\begin{table*}
    \centering
    \small
    \caption{
 The run time. Our methods: \texttt{$t$-FBIF},
 compared methods: $t$-CV (\texttt{$t$-CV}),
 $t$=5,10,20.}
 \label{tab-test-time}
    \begin{tabular*}{\linewidth}{@{\extracolsep{\fill}}crrrrrr}
   \toprule
    Classification     & \texttt{5-CV} &\texttt{5-FBIF}   &\texttt{10-CV}   &\texttt{10-FBIF}   &\texttt{20-CV} &\texttt{20-BIF}\\
    \midrule
       a1a         &36.80    &8.55    &87.50    &9.16    &251.46    &15.23\\
       a2a         &86.21    &16.11    &215.45    &20.46    &887.99    &30.82\\
       a3a         &155.01    &43.90    &254.52    &22.24    &1161.33    &30.12\\
       fourclass   &10.92    &3.14    &24.30    &3.01    &57.13    &4.01\\
       german     &10.47    &2.77    &22.69    &3.05    &48.57    &4.21\\
       madelon     &64.72    &14.57    &117.15    &14.13    &276.64    &15.31\\
       svmguide3   &15.82    &4.43    &25.93    &5.89    &56.29    &7.02\\
       splice   &11.43    &4.33    &30.92    &4.78    &64.53    &5.62\\
       w1a         &144.02    &22.69    &382.56    &23.47    &1188.36    &37.31\\
       w2a         &181.91    &26.04    &595.33    &30.89    &1147.22    &29.14\\
       w3a         &209.87    &29.40    &458.01    &23.88    &721.27    &17.75\\
       \midrule
       Regression     & \texttt{5-CV} &\texttt{5-FBIF}   &\texttt{10-CV}   &\texttt{10-FBIF}   &\texttt{20-CV} &\texttt{20-BIF}\\
       \midrule
       abalone  &101.09    &17.08    &195.75    &15.09    &217.93    &20.01\\
       bodfat   &1.10    &0.17    &2.50    &0.13    &5.59    &0.15\\
       cpusmall &212.43    &22.58    &481.85    &25.88    &1706.09 &37.61\\
      mg       &17.10    &5.74    &30.40    &6.46    &91.86    &11.27\\
       mpg      &1.58    &0.20    &4.20    &0.18    &9.20    &0.19\\
      housing  &1.29    &0.17    &3.18    &0.15    &7.25    &0.16\\
      space-ga &70.68    &20.58    &151.87    &18.58    &443.62    &16.68\\
    \bottomrule
 \end{tabular*}
 \vspace{-0.5cm}
\end{table*}
\subsection{Time Complexity}
To compute $t$-BIF, we need
$\mathcal{O}(l^3+(l+u)c^2+c^3)$ and $\mathcal{O}(l_\mathrm{sv}^3+c^3+(l+u)c^2)$
to compute the $\tilde{\mathbf H}^{-1}$ for LapRLS and LapSVM,
and $\mathcal{O}(l(l+u)+(l+u)c+t(l+u)c)$ to compute $\mathbf B$.
Since $f_{\mathbb{P}_\mathcal{S}}^\mathrm{Lap}$ has been obtained in the training process,
thus the overall time complexity of $t$-BIF for LapRLS is $\mathcal{O}(l^3+(l+u)c^2+c^3+t(l+u)c+l(l+u)),$
for LapSVM is $\mathcal{O}(l_\mathrm{cv}^3+(l+u)c^2+c^3+t(l+u)c+l(l+u)),$
which is much faster than the traditional $t$-CV of time complexity $\mathcal{O}\Big(\frac{(t-1)^3(l+u)^3}{t^2}\Big), l\ll u, c\ll u.$

\section{Experiment}
In this section,
we will compare our proposed approximate $t$-CV (\texttt{$t$-BIF}) with the original $t$-CV (\texttt{$t$-CV}), $t=5, 10,20$.
The data sets are 18 publicly available data sets from LIBSVM Data\footnote{http://www.csie.ntu.edu.tw/$\sim$cjlin/libsvm.}:
11 data sets for classification and 7 data sets for regression.
All data sets are normalized to
zero-mean and unit-variance on every attribute to
avoid numerical problems.
Experiments are performed on  a single machine with two cores (Intel
Xeon E5-2630@2.40 GHz) and 64 GB memory.
We use the Gaussian kernel
$\kappa(\mathbf  x,\mathbf  x')=\exp\left(-\frac{\|\mathbf  x-\mathbf  x'\|^2_2}{2\sigma}\right)$
as our candidate kernel
$\sigma\in \{2^i,i=-10,-8,\ldots,10\}$.
The candidate regularization  parameters
$
  \gamma_A\in\{10^i,i=-6,-5,\ldots, 2\},
\gamma_I\in\{10^i,i=-6,-5,\ldots, 2\}.
$
The candidate graph Laplacian parameters $k\in\{2,4,8\}$ and
$\sigma_w\in\{2^i, i=-4,-2,\ldots,4\}$.

The learning algorithm used in our experiments for regression is LapRLS and for classification is LapSVM.
For each data set, we run all methods 30 times with randomly selected $70\%$ of all data for
training and the other $30\%$ for testing.
Meanwhile, from each train data,
we randomly select 10\% examples as labeled data.
The use of multiple training/test partitions allows
an estimate of the statistical significance of differences
in performance between methods.
Let $A_i$ and $B_i$ be the test errors of methods A and B in partition $i$,
and $d_i = B_i-A_i$, $i=1,\ldots, 30$.
Let $\bar{d}$ and $S_d$ be the  mean and standard error of $d_i$.
Then
under $t$-test,
with confidence level $95\%$,
we claim  that A is significantly better than B (or equivalently B significantly worse than A)
if the $t$-statistic
$\frac{\bar{d}}{S_d/\sqrt{30}}> 1.699.$
All statements of statistical
significance in the remainder refer to a $95\%$ level of significance.
\subsection{Accuracy}
The test errors for classification and  test mean square errors for regression are
reported in Table \ref{tab-test-classification}.
For our \texttt{$t$-BIF}, we set $c=\sqrt{l+u}$ (in fact, we have tried many other setting of $c$ on some small datasets in advance, we find that if $c\geq \sqrt{l+u}$,
the accuracy of our approximate CV is good, so in this paper, we set $c=\sqrt{l+u}$ on all dataset for simplicity) and
set $h=0.01$ for LapSVM (note that if $h$ is small, the Huber loss is a good approximation of Hinge loss, thus we set $h=0.01$).
The elements are obtained as follows:
For each training set,
we select the kernel parameter $\sigma$, the regularization parameters $\gamma_A$ and $\gamma_I$, the  graph Laplacian parameters $k$ and $\sigma_w$,
by each  criterion on the training set,
and evaluate the test error for the chosen parameters on the test set.
The results in Table \ref{tab-test-classification} can be summarized as follows:
(1) Neither of \texttt{$t$-CV} and \texttt{$t$-BIF} for classification and regression
is statistically superior at the $95\%$ level of significance, $t$=5, 10, 20.
(2) For regression, \texttt{$t$-BIF} gives almost  the same testing errors as the the traditional $t$-CV, $t=5,10,20$.
In particular, on
abalone, bodyfat, cpusmall, mg, space-ga, \texttt{$t$-BIF}
gives the same testing errors as $t$-CV.
On the remaining data
sets, both \texttt{$t$-BIF} and $t$-CV give the similar results.

The above results implicate that the quality of our approximation based on
the BIF is quite good.

\subsection{Efficiency}
The run time of \texttt{$t$-BIF} and \texttt{$t$-CV} is  reported in Table \ref{tab-test-time}.
We can find that
\texttt{$t$-BIF} is  much faster  than \texttt{$t$-CV}.
    In particular, \texttt{$t$-BIF} is nearly $t$ (or more)  times
    faster than \texttt{$t$-CV} on most data sets.
    For large datasets, such as a3a, w1a, w2a, cpusmall, \texttt{$20$-BIF} is nearly 40 times faster than 20-CV.
    Thus, \texttt{$t$-BIF} significantly improves the efficiency of \texttt{$t$-CV} for model selection of manifold regularization.

\section{Conclusion}
In this paper,
we  present an approximate CV method
based on the  theoretical notion of BIF
for manifold regularization in semi-supervised learning.
The proposed approximate CV requires training on the full data only once,
hence can significantly improve the efficiency.
Experimental results on 18 data sets
show that our approximate CV much more efficiency and has no statistical discrepancy
when compared to the original one.
This is an interesting attempt to apply the theoretical notion of BIF
for practical model selection in semi-supervised learning.

Future work includes extending our results to other  manifold regularization algorithms,
such as square laplacian support vector machine and laplacian logistic regression.
\section*{Appendix: Proof of Theorem \ref{the-bif}}
\begin{proof}
  The derivative of the objective function vanishes at the minimizer, so we have
\begin{align}
  \label{derivative-origal}
  \begin{aligned}
  -2\gamma_A f^\mathrm{MR}_{\mathbb{P}_{\mathcal{S}}}=&\frac{(\mathbf J_\mathcal{S}\bm \phi_\mathcal{S})^\mathrm{T}\bm \mu_{\mathcal{S}}}{l}+
      \frac{2\gamma_I\bm \phi_\mathcal{S}^\mathrm{T} \mathbf L_{\mathcal{S}}\bm f_{\mathbb{P}_{\mathcal{S}}}^\mathcal{S}}{(l+u)^2},
  \end{aligned}
\end{align}
where $\bm \phi_{\mathcal{S}}=(\kappa(\mathbf x_1,\cdot),\ldots,\kappa(\mathbf x_{l+u},\cdot))^\mathrm{T}$,
$\bm \mu_{\mathcal{S}}$ is an $l+u$ vector with the first $l$ entries as $\ell'(y_{j},f^\mathrm{MR}_{\mathbb{P}_{\mathcal{S}}}(\mathbf x_{j}))$,
$(\mathbf x_j,y_j)\in \mathcal{L}$ and the rest $0$.

Denote $\mathbb{P}_{\mathcal{S}_{\epsilon,\mathcal{S}_i}}=(1-\epsilon)\mathbb{P}_{\mathcal{S}}+\epsilon \mathbb{P}_{\mathcal{S}_i}$,
we can obtain that
\begin{align}
\begin{aligned}
\label{equation-13}
  &-2\gamma_A f^\mathrm{MR}_{\mathbb{P}_{\mathcal{S}_{\epsilon,\mathcal{S}_i}}}
  \\=&\frac{(1-\epsilon)(\mathbf J_\mathcal{S}\bm \phi_\mathcal{S})^\mathrm{T}\bm \mu_{\mathcal{S},\epsilon}}{l}+
      \frac{2(1-\epsilon)\gamma_I\bm \phi_\mathcal{S}^\mathrm{T} \mathbf L_{\mathcal{S}}\bm f_{\mathbb{P}_{\mathcal{S}_{\epsilon,\mathcal{S}_i}}}^\mathcal{S}}{(l+u)^2}\\
      &+
       \frac{\epsilon(\mathbf J_{\mathcal{S}_i}\bm \phi_{\mathcal{S}_i})^\mathrm{T}\bm \mu_{\mathcal{S}_i,\epsilon}}{m}+
      \frac{2\epsilon\gamma_I\bm \phi_{\mathcal{S}_i}^\mathrm{T} \mathbf L_{\mathcal{S}_i}\bm f_{\mathbb{P}_{\mathcal{S}_{\epsilon,\mathcal{S}_i}}}^{\mathcal{S}_i}}{(m+n)^2},
  \end{aligned}
\end{align}
where $\bm \mu_{\mathcal{S},\epsilon}$ is an $l+u$ vector with the first $l$ entries as
 $\ell'(y_{j},f^\mathrm{MR}_{\mathbb{P}_{\mathcal{S}_{\epsilon,\mathcal{S}_i}}}(\mathbf x_{j}))$,
$(\mathbf x_j,y_j)\in \mathcal{L}$, and the rest $0$,
$\bm  f_{\mathbb{P}_{\mathcal{S}_{\epsilon,\mathcal{S}_i}}}^\mathcal{S}=\big(f^\mathrm{MR}_{\mathbb{P}_{\mathcal{S}_{\epsilon,\mathcal{S}_i}}}(\mathbf x_1),
\ldots, f^\mathrm{MR}_{\mathbb{P}_{\mathcal{S}_{\epsilon,\mathcal{S}_i}}}(\mathbf x_{l+u})\big)^\mathrm{T}$,
$\bm  f_{\mathbb{P}_{\mathcal{S}_{\epsilon,\mathcal{S}_i}}}^{\mathcal{S}_i}=\big(f^\mathrm{MR}_{\mathbb{P}_{\mathcal{S}_{\epsilon,\mathcal{S}_i}}}(\mathbf x^i_1),
\ldots, f^\mathrm{MR}_{\mathbb{P}_{\mathcal{S}_{\epsilon,\mathcal{S}_i}}}(\mathbf x^i_{m+n})\big)^\mathrm{T}$, where $\mathbf x_j^i\in\mathcal{T}_i$,
$\mathbf J_{\mathcal{S}_i}$ is an $(m+n)\times(m+n)$ diagonal matrix
with the first $m$ diagonal entries as $1$ and the rest $0$.
Taking the first derivative on both sides of (\ref{equation-13}) with respect to $\epsilon$ yields
\begin{align}
  \begin{aligned}
    \label{equation-14}
     &-2\gamma_A\frac{\partial}{\partial\epsilon} f^\mathrm{MR}_{\mathbb{P}_{\mathcal{S}_{\epsilon,\mathcal{S}_i}}}=\\
    &(1-\epsilon)\frac{1}{l}(\mathbf J_\mathcal{S}\bm \phi_\mathcal{S})^\mathrm{T}\mathbf F_{\mathcal{S},\epsilon}
    \left(\frac{\partial}{\partial \epsilon}
    \bm f_{\mathbb{P}_{\mathcal{S}_{\epsilon,\mathcal{S}_i}}}^{\mathcal{S}}\right)-\frac{1}{l}(\mathbf J_\mathcal{S}\bm \phi_\mathcal{S})^\mathrm{T}\bm \mu_{\mathcal{S},\epsilon}+\\
    &(1-\epsilon)\frac{2\gamma_I}{(l+u)^2}\bm \phi_{\mathcal{S}}^{\mathrm{T}} \mathbf L_{\mathcal{S}}\left(\frac{\partial}{\partial \epsilon}
    \bm f_{\mathbb{P}_{\mathcal{S}_{\epsilon,\mathcal{S}_i}}}^{\mathcal{S}}\right)-\frac{2\gamma_I}{(l+u)^2}
    \bm \phi_{\mathcal{S}}^{\mathrm{T}} \mathbf L_{\mathcal{S}}\bm f_{\mathbb{P}_{\mathcal{S}_{\epsilon,\mathcal{S}_i}}}^{\mathcal{S}}\\&
    +\frac{\epsilon}{m}(\mathbf J_{\mathcal{S}_i}\bm \phi_{\mathcal{S}_i})^\mathrm{T}\mathbf F_{\mathcal{S}_i,\epsilon}
    \left(\frac{\partial}{\partial \epsilon}
    \bm f_{\mathbb{P}_{\mathcal{S}_{\epsilon,\mathcal{S}_i}}}^{\mathcal{S}_i}\right)+\frac{1}{m}(\mathbf J_{\mathcal{S}_i}\bm \phi_{\mathcal{S}_i})^\mathrm{T}\bm \mu_{\mathcal{S}_i,\epsilon}+\\
    &\frac{2\epsilon\gamma_I}{(m+n)^2}\bm \phi_{\mathcal{S}_i}^{\mathrm{T}} \mathbf L_{\mathcal{S}_i}\left(\frac{\partial}{\partial \epsilon}
    \bm f_{\mathbb{P}_{\mathcal{S}_{\epsilon,\mathcal{S}_i}}}^{\mathcal{S}_i}\right)+\frac{2\gamma_I}{(m+n)^2}
    \bm \phi_{\mathcal{S}_i}^{\mathrm{T}} \mathbf L_{\mathcal{S}_i}\bm f_{\mathbb{P}_{\mathcal{S}_{\epsilon,\mathcal{S}_i}}}^{\mathcal{S}_i},
  \end{aligned}
\end{align}
where $\frac{\partial}{\partial \epsilon}
    \bm f_{\mathbb{P}_{\mathcal{S}_{\epsilon,\mathcal{S}_i}}}^{\mathcal{S}}=\big(\frac{\partial}{\partial \epsilon}
     f^\mathrm{MR}_{\mathbb{P}_{\mathcal{S}_{\epsilon,\mathcal{S}_i}}}(\mathbf x_1),\ldots,
    \frac{\partial}{\partial \epsilon}f^\mathrm{MR}_{\mathbb{P}_{\mathcal{S}_{\epsilon,\mathcal{S}_i}}}(\mathbf x_{l+u})\big)^\mathrm{T}$,
    $\frac{\partial}{\partial \epsilon}
\bm f_{\mathbb{P}_{\mathcal{S}_{\epsilon,\mathcal{S}_i}}}^{\mathcal{S}_i}=
    \big( \frac{\partial}{\partial \epsilon}f^\mathrm{MR}_{\mathbb{P}_{\mathcal{S}_{\epsilon,\mathcal{S}_i}}}(\mathbf x^i_1),\ldots,
    \frac{\partial}{\partial \epsilon}f^\mathrm{MR}_{\mathbb{P}_{\mathcal{S}_{\epsilon,\mathcal{S}_i}}}(\mathbf x^i_{m+n})\big)^\mathrm{T}$, $\mathbf x^i_j\in\mathcal{T}_i$,
    $\bm \phi_{\mathcal{S}_i}=(\kappa(\mathbf x^i_1,\cdot),\ldots,\kappa(\mathbf x^i_{m+n},\cdot))^\mathrm{T}$,
$\mathbf x^i_j\in\mathcal{T}_i$,
$\mathbf F_{\mathcal{S},\epsilon}$ is an $(l+u)\times(l+u)$ diagonal matrix
with the first $l$ diagonal entries as $\ell''(y_j,f^\mathrm{MR}_{\mathbb{P}_{\mathcal{S}_{\epsilon,\mathcal{S}_i}}}(\mathbf x_j))$ and the rest $0$,
$\mathbf F_{\mathcal{S}_i,\epsilon}$ is an $(m+n)\times(m+n)$ diagonal matrix with
the first $m$ diagonal entries as $\ell''(y^i_j,f^\mathrm{MR}_{\mathbb{P}_{\mathcal{S}_{\epsilon,\mathcal{S}_i}}}(\mathbf x^i_j))$,
$(\mathbf x_j^i,y_j^i)\in \mathcal{L}_i$, and the rest $0$,
$\bm \mu_{\mathcal{S}_i,\epsilon}$ is an $m+n$ vector with the first $m$ entries as
 $\ell'(y_{j}^i,f^\mathrm{MR}_{\mathbb{P}_{\mathcal{S}_{\epsilon,\mathcal{S}_i}}}(\mathbf x_{j}^i))$,
$(\mathbf x^i_j,y^i_j)\in \mathcal{L}_i$, and the rest $0$.

Setting $\epsilon=0$ on \eqref{equation-14}, we have $-2\gamma_A\frac{\partial}{\partial\epsilon} f^{\mathrm{MR}}_{\mathbb{P}_{\mathcal{S}_{\epsilon,\mathcal{S}_i}}}\big|_{\epsilon=0}=$
\begin{align}
  \begin{aligned}
    \label{equation-15}
    &\frac{1}{l}(\mathbf J_\mathcal{S}\bm \phi_\mathcal{S})^\mathrm{T}\mathbf F_{\mathcal{S}}
    \left(\frac{\partial}{\partial \epsilon}
    \bm f_{\mathbb{P}_{\mathcal{S}_{\epsilon,\mathcal{S}_i}}}^{\mathcal{S}}\big|_{\epsilon=0}\right)
    -\frac{1}{l}(\mathbf J_\mathcal{S}\bm \phi_\mathcal{S})^\mathrm{T}\bm \mu_{\mathcal{S}}\\
    &+\frac{2\gamma_I}{(l+u)^2}\bm \phi_{\mathcal{S}}^{\mathrm{T}} \mathbf L_{\mathcal{S}}\left(\frac{\partial}{\partial \epsilon}
    \bm f_{\mathbb{P}_{\mathcal{S}_{\epsilon,\mathcal{S}_i}}}^{\mathcal{S}}\big|_{\epsilon=0}\right)-\frac{2\gamma_I}{(l+u)^2}
    \bm \phi_{\mathcal{S}}^{\mathrm{T}} \mathbf L_{\mathcal{S}}\bm f_{\mathbb{P}_{\mathcal{S}}}^{\mathcal{S}}\\&+
    \frac{1}{m}(\mathbf J_{\mathcal{S}_i}\bm \phi_{\mathcal{S}_i})^\mathrm{T}\bm \mu_{\mathcal{S}_i}+\frac{2\gamma_I}{(m+n)^2}
    \bm \phi_{\mathcal{S}_i}^{\mathrm{T}} \mathbf L_{\mathcal{S}_i}\bm f_{\mathbb{P}_{\mathcal{S}}}^{\mathcal{S}_i},
  \end{aligned}
\end{align}
where
$
\frac{\partial}{\partial \epsilon}
    \bm f_{\mathbb{P}_{\mathcal{S}_{\epsilon,\mathcal{S}_i}}}^{\mathcal{S}}\big|_{\epsilon=0}\\
    =\big(\frac{\partial}{\partial \epsilon}
     f_{\mathbb{P}_{\mathcal{S}_{\epsilon,\mathcal{S}_i}}}^{\mathrm{MR}}\big|_{\epsilon=0}(\mathbf x_1),\ldots,
     \frac{\partial}{\partial \epsilon}
     f_{\mathbb{P}_{\mathcal{S}_{\epsilon,\mathcal{S}_i}}}^{\mathrm{MR}}\big|_{\epsilon=0}(\mathbf x_{l+u})\big).
$
According to \eqref{derivative-origal}, we know that
$
  -2\gamma_A f^\mathrm{MR}_{\mathbb{P}_{\mathcal{S}}}=\frac{(\mathbf J_\mathcal{S}\bm \phi_\mathcal{S})^\mathrm{T}\bm \mu_{\mathcal{S}}}{l}+
      \frac{2\gamma_I}{(l+u)^2}\bm \phi_\mathcal{S}^\mathrm{T} \mathbf L_{\mathcal{S}}\bm f_{\mathbb{P}_{\mathcal{S}}}^\mathcal{S}.
$
Substituting the above Equation into \eqref{equation-15},
we have
\begin{align}
  \begin{aligned}
    \label{equation-16}
     &-2\gamma_A\frac{\partial}{\partial\epsilon} f^\mathrm{MR}_{\mathbb{P}_{\mathcal{S}_{\epsilon,\mathcal{S}_i}}}\big|_{\epsilon=0}=\frac{1}{l}(\mathbf J_\mathcal{S}\bm \phi_\mathcal{S})^\mathrm{T}\mathbf F_{\mathcal{S}}
    \left(\frac{\partial}{\partial \epsilon}
    \bm f_{\mathbb{P}_{\mathcal{S}_{\epsilon,\mathcal{S}_i}}}^{\mathcal{S}}\big|_{\epsilon=0}\right)\\
    &+\frac{2\gamma_I}{(l+u)^2}\bm \phi_{\mathcal{S}}^{\mathrm{T}} \mathbf L_{\mathcal{S}}\left(\frac{\partial}{\partial \epsilon}
    \bm f_{\mathbb{P}_{\mathcal{S}_{\epsilon,\mathcal{S}_i}}}^{\mathcal{S}}\big|_{\epsilon=0}\right)+2\gamma_A f_{\mathbb{P}_\mathcal{S}}\\&+
    \frac{1}{m}(\mathbf J_{\mathcal{S}_i}\bm \phi_{\mathcal{S}_i})^\mathrm{T}\bm \mu_{\mathcal{S}_i}+\frac{2\gamma_I}{(m+n)^2}
    \bm \phi_{\mathcal{S}_i}^{\mathrm{T}} \mathbf L_{\mathcal{S}_i}\bm f_{\mathbb{P}_{\mathcal{S}}}^{\mathcal{S}_i}.
  \end{aligned}
\end{align}
From equation \eqref{equation-16},
it is easy to verity that
\begin{align}
\begin{aligned}
  \label{BIFcol}
  &-2\gamma_A\left(\frac{\partial}{\partial \epsilon}
    \bm f_{\mathbb{P}_{\mathcal{S}_{\epsilon,\mathcal{S}_i}}}^{\mathcal{S}}\big|_{\epsilon=0}\right)=\frac{1}{l}\mathbf J_\mathcal{S}\mathbf K_{\mathcal{S}\mathcal{S}}\mathbf F_{\mathcal{S}}
    \left(\frac{\partial}{\partial \epsilon}\bm f_{\mathbb{P}_{\mathcal{S}_{\epsilon,\mathcal{S}_i}}}^{\mathcal{S}}\big|_{\epsilon=0}\right)\\
  &+\frac{2\gamma_I}{(l+u)^2}\mathbf K_{\mathcal{S}\mathcal{S}}\mathbf L_{\mathcal{S}}\left(\frac{\partial}{\partial \epsilon}
    \bm f_{\mathbb{P}_{\mathcal{S}_{\epsilon,\mathcal{S}_i}}}^{\mathcal{S}}\big|_{\epsilon=0}\right)+2\gamma_A\bm f_{\mathbb{P}_{\mathcal{S}}}^{\mathcal{S}}
  \\&+\frac{1}{m}\mathbf K_{\mathcal{S}\mathcal{S}_i}\bm\mu_{\mathcal{S}_i}
  +\frac{2\gamma_I}{(m+n)^2}\mathbf K_{\mathcal{S}\mathcal{S}_i}\mathbf L_{\mathcal{S}_i}\bm f^{\mathcal{S}_i}_{\mathbb{P}_{\mathcal{S}}}.
\end{aligned}
\end{align}
Since $\mathrm{BIF}(\mathbb{P}_{\mathcal{S}_i};f^{\mathrm{MR}},\mathbb{P}_\mathcal{S})$ is the  first order derivative
of $f^{\mathrm{MR}}_{\mathbb{P}_{\mathcal{S}_{\epsilon,\mathcal{S}_i}}}$ at
$\epsilon=0$,
we have
$
  \mathbf B_{\cdot,i}=\left(\frac{\partial}{\partial \epsilon}\bm f_{\mathbb{P}_{\mathcal{S}_{\epsilon,\mathcal{S}_i}}}^{\mathcal{S}}\big|_{\epsilon=0}\right).
$
Substituting the above Equation into \eqref{BIFcol}, which finishes the proof.
\end{proof}
\bibliographystyle{named}
\bibliography{ijcai19}

\end{document}